%% file: paper.tex
\pdfoutput=1
\documentclass[11pt]{article}

\usepackage{ACL2023}

\usepackage{balance}
\usepackage{times}
\usepackage{latexsym}
\usepackage{graphicx}
\usepackage{makecell}
\usepackage{longtable}
\usepackage{multirow}
\usepackage[normalem]{ulem}
\useunder{\uline}{\ul}{}
\usepackage[T1]{fontenc}
\usepackage[utf8]{inputenc}
\usepackage{microtype}
\usepackage{inconsolata}
\usepackage{multirow}
\usepackage{pdflscape}

\title{Hierarchical Evaluation Framework: Best Practices for Human Evaluation}

\author{Iva Bojic$^{1}$ \and Jessica Chen$^{2}$ \and Si Yuan Chang$^{1}$ \and Qi Chwen Ong$^{1}$ \and \\ {\bf Shafiq Joty$^{1,3}$ \and Josip Car$^{1,2}$} \\
$^1${Nanyang Technological University Singapore} \\
$^2${Imperial College London, United Kingdom} \\
$^3${Salesforce Research, USA} 
}

\begin{document}
\maketitle
\begin{abstract}
Human evaluation plays a crucial role in Natural Language Processing (NLP) as it assesses the quality and relevance of developed systems, thereby facilitating their enhancement. However, the absence of widely accepted human evaluation metrics in NLP hampers fair comparisons among different systems and the establishment of universal assessment standards. Through an extensive analysis of existing literature on human evaluation metrics, we identified several gaps in NLP evaluation methodologies. These gaps served as motivation for developing our own hierarchical evaluation framework. The proposed framework offers notable advantages, particularly in providing a more comprehensive representation of the NLP system's performance. We applied this framework to evaluate the developed Machine Reading Comprehension system, which was utilized within a human-AI symbiosis model. The results highlighted the associations between the quality of inputs and outputs, underscoring the necessity to evaluate both components rather than solely focusing on outputs. In future work, we will investigate the potential time-saving benefits of our proposed framework for evaluators assessing NLP systems.
\end{abstract}

\section{Introduction}
\input{Sections/1_intro}

\section{Scoping Review}
\label{rw}
\input{Sections/2_rw}

\section{Hierarchical Evaluation Framework}
\label{hierarchical}
\input{Sections/3_hierarchical}

\section{Case study: Hierarchical Evaluation for an MRC System}
\label{case}
\input{Sections/4_case}

\section{Discussion and Conclusions}
\label{dis}
\input{Sections/5_dis}

\section*{Limitations}
We recognize the potential limitations that may arise with a small-scale scoping review that is limited to a few venues. As our sample size is small, our results and proposed solutions may lack generalizability and applicability. To mitigate the potentially negative effects, we carefully chose the most appropriate venues - as further explained in \ref{sec:Structured review} - and limited the search to the most recent papers as the field of computer science is rapidly and constantly evolving. Solely reviewing papers in the English language could also potentially limit the scope of our research. We also tried to delve into a broad range of aspects of human evaluation whilst keeping our objectives focused. However, we recognize the inevitability of potential factors that may exist outside of our considerations - which may also affect results and conclusions.

\section*{Ethics Statement}
We aim to conduct our study with the highest ethical standards and maintain continuous referral to the ACL code of ethics throughout our research. We obtained articles via Google Scholar and have anonymized most of the papers and authors - excluding a few that were cited in our main text. This paper should be used to provide insight into the current practices of human evaluation and a potential solution to streamline the process. It is not used to penalize any research or draw any negative attention to certain papers. 

We also recognize that some potential biases and errors may arise amongst human reviewers which may lead to potentially inaccurate data extraction. This may have a potential knock-on effect on derived conclusions. These issues are considered and mitigated through multiple reviewers performing the same task, frequent discussions, and good communication.

\section*{Acknowledgements}

The authors would like to acknowledge the Accelerating Creativity and Excellence (ACE) Award (NTU-ACE2020-05) and center funding from Nanyang Technological University, Singapore. Josip Car’s post at Imperial College London is supported by the NIHR NW London Applied Research Collaboration. Finally, the authors would also like to acknowledge Jintana Liu and Ashwini Lawate who were included in the pilot RCT running and supported the data collection process.

\balance
\bibliography{references}
\bibliographystyle{acl_natbib}

\section*{Appendix}
\label{app}

\subsection*{Appendix 1: Data Extraction Form}
    \begin{figure*}[t!]
        \centering  \includegraphics[height=1.55\linewidth]{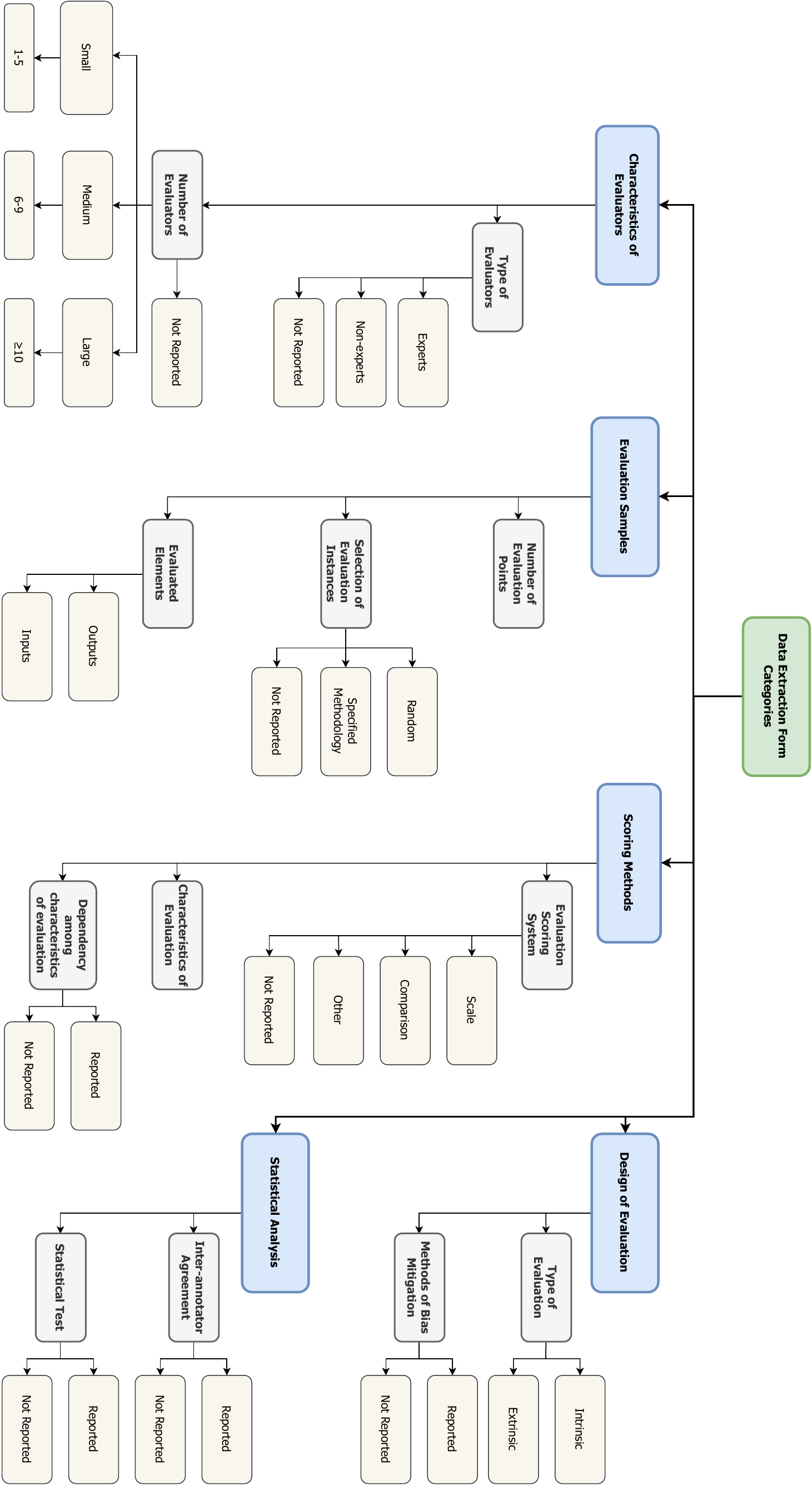}
        \caption{Data extraction form categories.}
        \label{fig:Data Extraction Form Categories}
    \end{figure*}

\end{document}

%% file: Sections/1_intro.tex
Human evaluation is crucial for assessing the quality, validity, and performance of Natural Language Processing (NLP) systems especially as automatic metrics are usually not sufficient \cite{van2019best}. Human evaluation can deal with complex generated natural language and its nuances such as pragmatics, context and semantics which often requires some expert knowledge \cite{sudoh2021translation}. Automatic evaluation may be used to assess individual dimensions (e.g., fluency, accuracy) of natural language, however, may often lose to humans in terms of accuracy and understanding. 

Various methodologies are often employed in human evaluation such as ranking, pairwise comparison, or a state-of-the-art machine translation metric that was used in \citet{castilho2021towards}. They can provide valuable insights into the strengths and limitations of an NLP system; however, it is notably time-consuming and expensive and significant trade-offs may exist in consideration of different goals or requirements \cite{zhang2020trading}. The human evaluation also comes with its own set of limitations, such as fatigue effect \cite{van2021human} and inconsistencies between evaluators. The role of human evaluators should also be considered as some tasks may require domain expert knowledge or provide specific training evaluators.  

There is currently a lack of consensus on which metrics to use for the human evaluation of NLP systems \cite{paroubek-etal-2007-principles}. As there tend to be different research goals, requirements and task-dependent metrics, there exists the challenge of standardizing human evaluation metrics and essentially reaching an overall consensus. A unique combination of metrics can be used for a more comprehensive assessment depending on the desired objectives. These combinations can be grouped based on different evaluation aspects \cite{liang2021towards}. Metrics may also vary depending on the task (e.g., machine translation, sentiment analysis) and thus task design can affect the criteria used for evaluation \cite{iskender2021reliability}. 

To identify gaps in the literature pertaining to human evaluation, we conducted a scoping review to systematically examine various aspects of human evaluation experiments in NLP tasks, including the characteristics of evaluators, evaluation samples, scoring methods, design of evaluation and statistical analysis. The findings of our literature review revealed three significant gaps: (i) the absence of evaluation metrics for NLP system inputs, (ii) the lack of consideration for interdependencies among different characteristics of assessed NLP systems, and (iii) a limited utilization of metrics for extrinsic evaluation of NLP systems.

We hope to bridge the aforementioned gaps by providing a standardized human evaluation framework that can be used across different NLP tasks. Our proposed framework employs a hierarchical structure that divides the human evaluation process into two phases: testing and evaluation. This division enables evaluators to assess the quality of inputs used by testers when evaluating NLP systems. Furthermore, the hierarchical design of the evaluation metric allows for the computation of a composite score that reflects the overall quality of the NLP system.

This paper is organized as follows. Section \ref{rw} presents the analysis from a scoping review that included more than 200 papers published within the last three years in the top 5 NLP venues. The results of the aforementioned analysis informed the development of the proposed hierarchical evaluation framework, which is presented in Section \ref{hierarchical}. Section \ref{case} presents the results of adopting the proposed framework for the human evaluation of the Machine Reading Comprehension (MRC) system developed as a part of the human-AI symbiosis model. Finally, Section \ref{dis} concludes the paper.

%% file: Sections/2_rw.tex
\subsection{Structured Review}
\label{sec:Structured review}
\begin{figure}[h!]
\centering 
\includegraphics[width=0.935\linewidth]{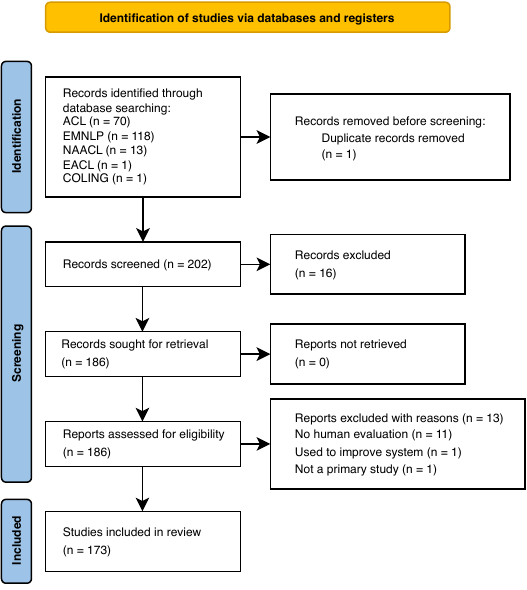}
\caption{This PRISMA flow diagram depicts the study selection process throughout this scoping review. 203 studies in total were identified through a search on Google Scholar. After one duplicate was removed, the total remaining studies was 202. After title and abstract screening, 16 studies were excluded, leaving 186 studies for full-text screening. A final 173 studies were included in this scoping review for data extraction and analysis.}
    \label{PRISMA}
\end{figure}

To inform our development of a hierarchical framework for human evaluation, we conducted a scoping review to examine existing literature systematically. Our paper selection process followed the Preferred Reporting Items for Systematic Reviews and Meta-Analyses (PRISMA) extension for Scoping Reviews checklist (PRISMA-ScR) \cite{peters2015joanna} (see Figure~\ref{PRISMA}). We searched for relevant publication venues on Google Scholar. We selected the category of Engineering and Computer Science, followed by the sub-category of Computational Linguistics. Subsequently, we chose the top five venues with the highest h5-index, namely:

\begin{itemize}
    \item Meeting of the Association for Computational Linguistics (ACL),
    \item Conference on Empirical Methods in Natural Language Processing (EMNLP),
    \item Conference of the North American Chapter of the Association for Computational Linguistics: Human Language Technologies (NAACL),
    \item Conference of the European Chapter of the Association for Computational Linguistics (EACL),
    \item International Conference on Computational Linguistics (COLING). 
\end{itemize}

\begin{figure}[h!]
\centering 
\includegraphics[width=1\linewidth]{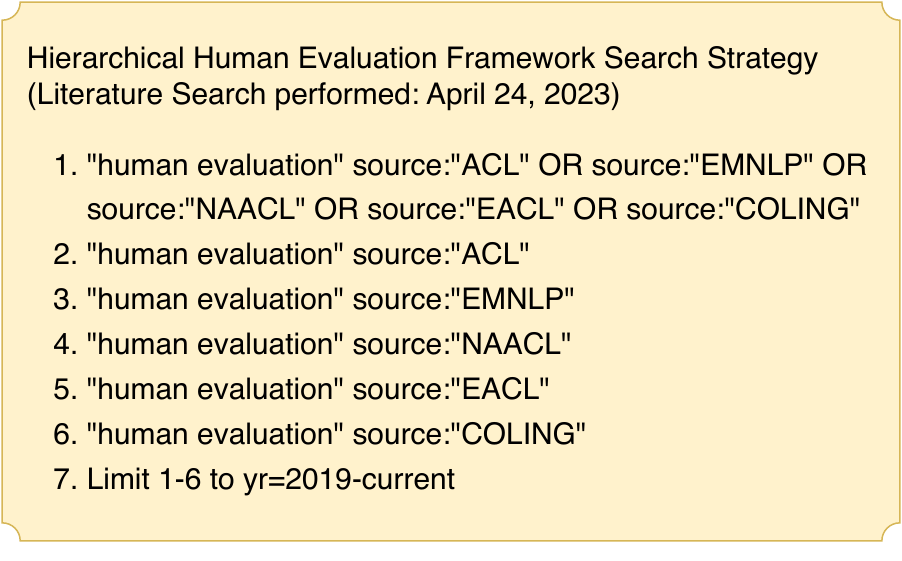}
\caption{Search strategy used for the scoping review. After performing 1, we also performed 2-6 to find all papers from individual venues that did not appear after the first combined search.}
    \label{Google Scholar Search Strategy}
\end{figure}

Due to the rapid development in the NLP field, only studies published between 2019 and 2023 were included. The Google Scholar search strategy is shown in Figure \ref{Google Scholar Search Strategy}. 

\subsection{Selection of Articles}

\begin{figure}[h!]
\centering 
\includegraphics[width=0.935\linewidth]{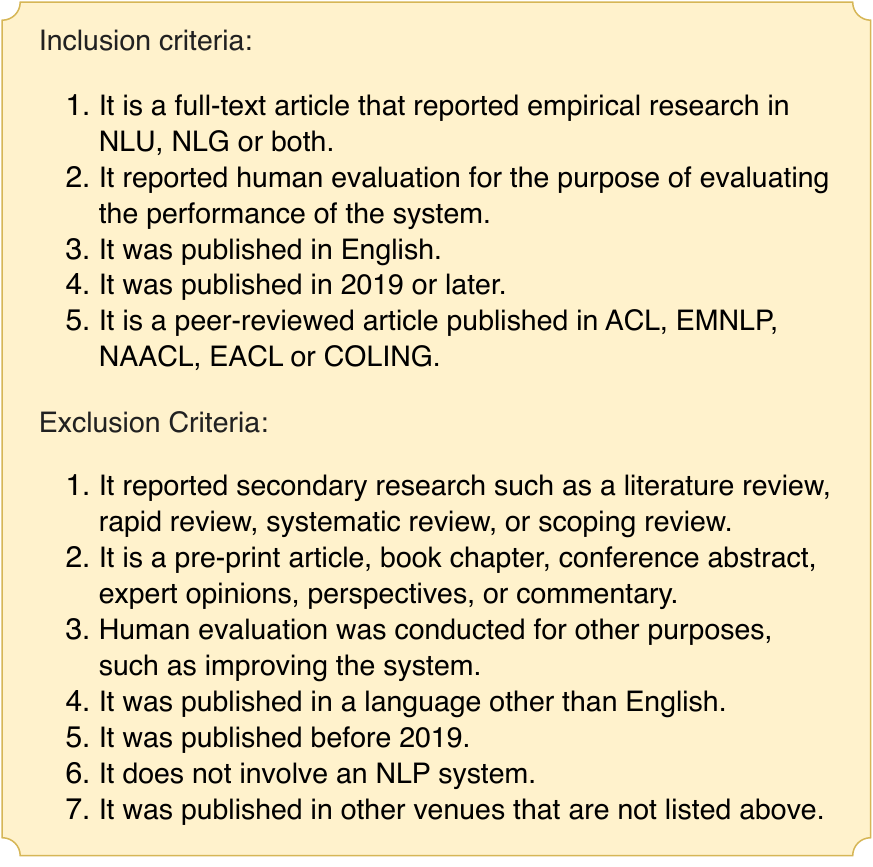}
\caption{This figure lists the inclusion and exclusion criteria that formed the basis of our screening process.}
    \label{Inclusion and Exclusion Criteria}
\end{figure}

Eligible articles were identified in two stages: (1) title and abstract screening, (2) full-text screening. To maintain consistency of decision-making in the selection process, both title and abstract screening and full-text screening were conducted by two of the three reviewers (IB, JC, QCO) independently based on pre-defined inclusion and exclusion criteria (see Figure \ref{Inclusion and Exclusion Criteria}). Conflicts were resolved through discussion with a third reviewer to establish consensus. The resolution of inconsistencies or disagreements amongst reviewers was guided by pre-defined eligibility criteria and reference to initial objectives. Reasons for exclusion were recorded during full-text screening.

\subsection{Data Extraction}

A standardized data extraction form (see Appendix~1) was developed through iterative discussions between three reviewers (IB, JC, QCO) based on insights gained during the initial literature review of related work. The data extraction form was first piloted on three randomly selected articles by the three reviewers to ensure consistent and accurate extraction of data. The data extraction process involved all three reviewers and was done independently. Ambiguities or uncertainties were resolved by discussion between reviewers and by referring to the original papers used for the creation of the extraction matrix \cite{van2019best, amidei2018evaluation, liang2021towards, howcroft2020twenty}. We extracted a range of variables from certain chosen sources and tailored them to the objectives of our review. These variables are categorized as follows in Section \ref{sec:Synthesis of results}: (1) characteristics of evaluators, (2) evaluation samples, (3) scoring methods, (4) design of evaluation and (5) statistical analysis.

\subsection{Synthesis of Results}
\label{sec:Synthesis of results}

\subsubsection{Characteristics of Evaluators} 

A large proportion of papers (83\%, 144/173) provided information on the number of evaluators that participated in the human evaluation. This shows that there is a general consistency in the reporting of human evaluation methods across all papers reviewed. The number of evaluators employed can be defined as \textit{small} (1-5), \textit{medium} (6-9) and \textit{large} ($\geq 10$) scale \cite{van2021human}. Papers reported a small number of evaluators in 62\% of cases (107/173), a medium number in 6\% (11/173), and a large number in 15\% (26/173). The median number of evaluators was three per study. 

71\% of the reviewed papers (122/173) reported the background of the evaluators, differentiating between \textit{experts} and \textit{non-experts}, detailed which platform they were from or set standards for crowd-sourced workers. One example, proposed in \citet{zhu2020counterfactual}, was to set standards by only using workers with a high enough approval rate to ensure quality. This helps alleviate the problem of quality control when using larger-scale crowd-sourcing platforms such as Amazon Mechanical Turk.

\subsubsection{Evaluation Samples} 

All of the papers reported that human evaluation was done only on \textit{outputs} of NLP systems, with the median number of evaluation instances being 100. Most papers (60\%, 103/173) created samples \textit{randomly}, but some (3\%, 6/173) specified \textit{their methodology}. For instance, in \citet{zeng2021investigation}, discussions that were difficult to understand were filtered out. In this case, human evaluation was used to compare the dialogue generation between two different models. In order to create a more relevant dataset for human evaluation, filtering out professional texts that were difficult to understand, ensured that the data was closer to daily dialogue. This allowed for more accurate and reproducible human evaluation results. Using alternative methods to random sampling can have certain benefits such as cost-effectiveness, time efficiency and focused research objectives \cite{zeng2021investigation}.

\subsubsection{Scoring Methods} 
\label{sec:Scoring Systems}

Overall, 68\% of papers (118/173) used a \textit{scale} as their evaluation scoring system. A scoring system should also be defined by assigning attributes or certain qualities to a number in the scale that they are using. Further, 23\% of papers (39/173) reported using \textit{comparison} between different models or question answering to achieve more qualitative results. Examples include win, tie, loss, A/B testing, and a direct comparison.

The characteristics of evaluation can be referred to as evaluation attributes or text quality dimensions such as \textit{fluency}, \textit{adequacy}, and \textit{grammar} \cite{gehrmann2023repairing}. These characteristics can be considered for both qualitative and quantitative methods and are often specified to guide the evaluation task. For example, \citet{liang2021towards} divided various characteristics into seven groups based on their similarity and overall purpose for the human evaluation of chatbots. These groups further tailor the characteristics of evaluation to the unique task, allowing the reader to understand the reason for their selection.

Dependencies can exist among characteristics of evaluation. In other words, human evaluation can be done in sequential order when the order in which characteristics are evaluated matters. Moreover, evaluation can be prematurely stopped if some characteristics were not deemed of a satisfactory quality. Consequently, dependencies among characteristics of evaluation could also allow for a NLP system to have a composite score that would reflect its overall quality. For instance, an overall performance score can be produced based on pre-defined threshold criteria that need to be fulfilled. This threshold could be a specified performance level reached by a specific combination of characteristics. We have not observed any dependencies reported among different evaluated characteristics in the reviewed literature. Namely, all characteristics were evaluated separately, and the quality of a certain characteristic was never put in relation with the quality of another one.

\subsubsection{Design of Evaluation} 

\textit{Extrinsic} and \textit{intrinsic} evaluation are two different types of human evaluation. Extrinsic evaluation assesses the ability of the system to perform an over-arching task with a real-world application. On the other hand, intrinsic evaluation assesses specific qualities or attributes and is evaluated independently of the over-arching task. Therefore, a system could perform well intrinsically without performing well extrinsically. Most papers (88\%, 153/173) performed intrinsic evaluation, 4\% (7/173) performed extrinsic evaluation, and 8\% (13/173) involved aspects of both intrinsic and extrinsic evaluation. Intrinsic evaluation remains popular likely due to its simplicity, cost-efficiency, ease in tracking progress and benchmarking \cite{gehrmann2023repairing}, \cite{belz2008intrinsic}. The lack of extrinsic evaluation may also be affected by the difficulty of designing an evaluation that effectively emulates its usage in the real-world setting. 

Bias mitigation is important due to the potential compromise of human evaluation caused by order effects \cite{van2019best}. Order effects include practice, carryover, and fatigue effects \cite{van2019best}, all of which have the potential to affect human evaluation and lead to misleading and biased results. To mitigate this, \citet{van2019best} suggested potential solutions including practice trials, increasing the time between tasks, shortening tasks, and proposed specific evaluation designs such as counterbalancing (systematically varying the order of presentation) and randomization. Further solutions include multiple evaluators assessing the same point \cite{son2022translating} to increase the reliability of their human evaluation and randomized counterbalancing, which is a combination of randomization and counterbalancing methods \cite{kurisinkel2019set}. However, the method of bias mitigation was only specified in 14\% (24/173) of papers. This may be due to the high costs of evaluation designs, specifically counterbalancing. However, according to \citet{van2019best}, randomization or limiting the evaluation to one judge per system (if order effects are suspected) should be sufficient to mitigate order effects and avoid biased results.

\subsubsection{Statistical Analysis} 

Inter-annotator agreement (IAA) scores should be reported to confirm consistency between evaluators and the reliability of the evaluation. Typically, a higher score indicates increased IAA. 34\% of included papers (58/173) reported IAA using Kendall’s $\tau$, Fleiss' $\kappa$, Cohen's $\kappa$, Krippendorf's $\alpha$ and percentage agreement to name a few. However, a detailed analysis of the IAA scores and how they affected the overall evaluation is important. In some cases, IAA scores can prove to not be a useful measurement of agreement - as alluded to further in \cite{amidei2018rethinking}.  

The importance of ensuring the reliability and validity of human evaluation is further highlighted by \citet{liu2022revisiting} through the need for using statistical tests. 
%The authors of the paper also alluded that larger datasets enable more statistically significant and stable results and emphasized the need for statistical analysis of evaluation of large language models in particular to identify potential biases present. 
Other methods of presenting data and analyzing results include displaying 1st and 2nd best performances in a table by highlighting the specific performance values \cite{gangal2022nareor}; or summary statistics such as standard deviations or mean scores \cite{qian2022flexible}. Only 16\% of papers (28/173) used statistical tests as a form of analysis of their human evaluation such as student's t-test and Wilcoxon ranked test \cite{van2019best}. This could be due to a lack of statistical power attributed to inadequate sample sizes, which could lead to misleading or different conclusions as they are more subject to the effects of chance \cite{otani2023toward}.

%% file: Sections/3_hierarchical.tex
The review of existing literature identified 3 gaps:

\begin{itemize}
    \item Majority of human evaluation was \textit{intrinsic}. 
    \item The characteristics of NLP systems were evaluated \textit{independently}.
    \item Human evaluation focused on assessing the \textit{outputs} of NLP systems, neglecting the evaluation of their \textit{inputs}. 
\end{itemize}

The analysis of existing literature revealed that the majority of papers (88\%, 153/173) focused solely on an intrinsic evaluation of NLP systems. To avoid conducting an evaluation merely for the sake of it, we suggest that first a clear purpose for an NLP system is defined, and subsequently, an extrinsic evaluation is designed to gauge the systems' performance in fulfilling that specific purpose.

Additionally, the evaluation of various aspects of NLP systems' outputs (e.g., truthfulness) is usually conducted independently, without providing a composite score for the overall system performance. We suggest adopting a hierarchical approach, where the characteristics of the systems are interdependent, and the evaluation process continues only if the preceding characteristic(s) is deemed satisfactory. Conversely, if a characteristic is unsatisfactory, the evaluation can be discontinued, allowing evaluators to save time by not evaluating all characteristics for the low-quality outputs.

Lastly, to date, the existing literature has focused solely on the human evaluation of NLP systems' outputs, assuming that the inputs provided to these systems were of good quality. However, this assumption may not always hold true. We thus propose a two-phase approach for human evaluation, wherein testers initially assess NLP systems, followed by evaluators who evaluate both the inputs and outputs of the systems. By dividing the evaluation process into two phases, we enable evaluators to also assess the quality of the inputs used by testers during the testing phase of NLP systems. In essence, our hypothesis is that the quality of the outputs may not only be influenced by the system itself but also by the quality of the inputs.

\begin{figure}[ht!]
\centering
\includegraphics[width=1\linewidth]{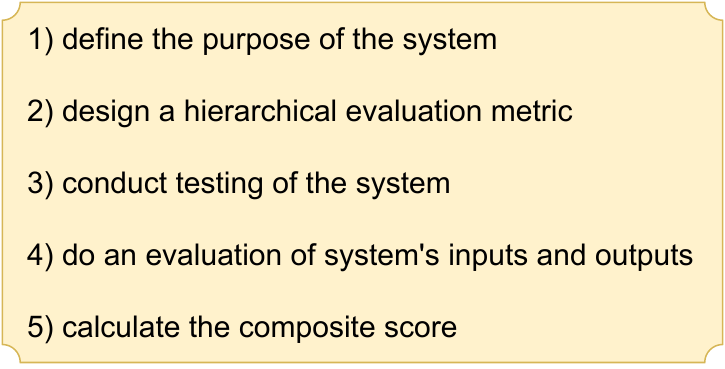}
\caption{Steps explaining how to create a hierarchical evaluation framework for an NLP system.}
\label{fig:framework}
\end{figure}

In order to address those gaps, we propose a framework as shown in Figure \ref{fig:framework}. By defining a system's purpose as the first step, our framework supports extrinsic evaluation. The second step is to define interdependencies between the evaluated characteristics and consequently to design a hierarchical evaluation metric that supports calculating a composite score that encompasses the overall quality of an NLP system. Namely, the evaluation stops if any of the evaluated characteristics is deemed unsatisfactory and, in this case, the composite score is ``bad'' as the system did not pass the evaluation. Otherwise, if the evaluation goes to the end, then the composite score is ``good''. We hypothesize that our framework facilitates a shorter evaluation time for evaluators by allowing early termination of evaluation in cases where any evaluated characteristic does not meet satisfactory quality. The third step is to do testing of the system according to the defined purpose. Testers are independent of evaluators who evaluate the system's inputs and outputs using the designed hierarchical evaluation metric in the fourth step. This allows for independent evaluation of the system's inputs as well. Consequently, our framework enables an examination of whether the quality of a system's outputs is influenced by the quality of its inputs. 

%% file: Sections/4_case.tex
We evaluated a Machine Reading Comprehension (MRC) system using the framework outlined in the previous section. In an MRC system, answers come in the form of short text spans which are directly extracted from the text corpus (i.e., relevant text database). Questions asked, on the other hand, need to be relevant to the topic that the text corpus covers, factoid, answerable and mistake-free (i.e., no spelling or grammar mistakes).

\subsection{The purpose of the MRC System}

\begin{figure}[ht!]
\centering
\includegraphics[width=1\linewidth]{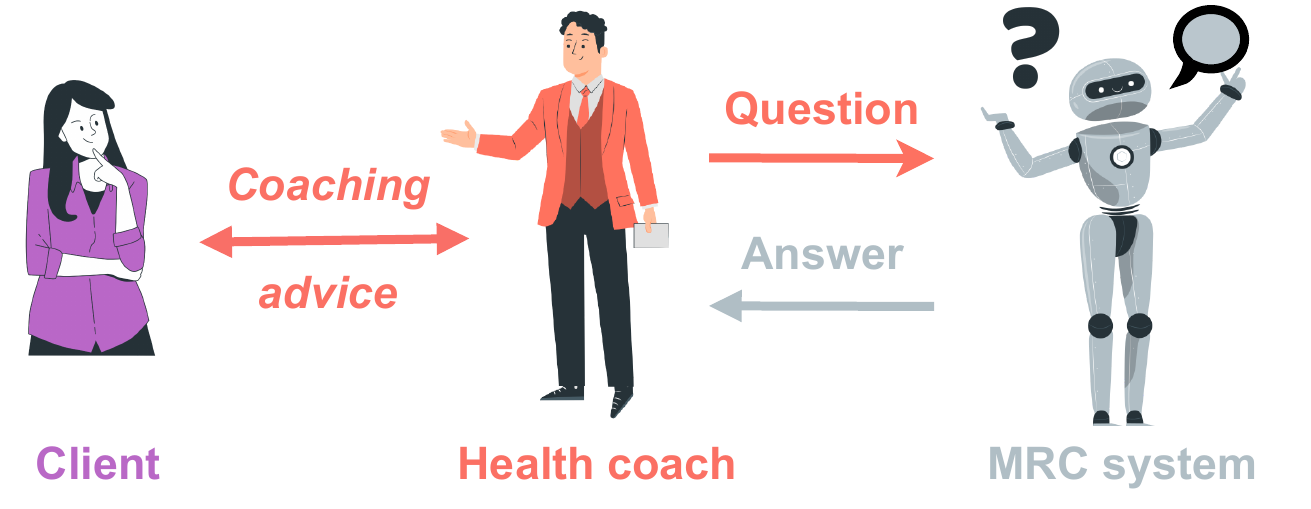}
\caption{Human-AI health coaching model.}
\label{fig:model}
\end{figure}

The purpose of the developed MRC system was to support health coaches during their sessions with clients, coaching them on the importance of good quality sleep. Namely, the developed system is part of the human-AI symbiosis model shown in Figure \ref{fig:model} \cite{bojic2023building}. The system is a pre-trained BERT model that was fine-tuned on a human-annotated domain-specific dataset. 

The entire health coaching process takes place online through text messaging. To address factoid questions raised by clients, the health coach may utilize the MRC system for additional support during coaching sessions \cite{bojic2022sleepqa, bojic-etal-2023-data}. Health coaches were given the liberty to use, modify, or disregard the answers provided by the MRC system. This integration enhances the human coaching experience by incorporating evidence-based knowledge given by the MRC system. As a result, the health coaches' response time improves, and the information they offer is grounded in reliable evidence.

\subsection{Hierarchical Evaluation Metrics}
\label{metric}

We developed two evaluation metrics: one for the inputs (i.e., questions) of the MRC system and the other for the outputs (i.e., answers), in order to be able to detect whether the quality of the MRC system output is affected by the quality of its input. 

\subsubsection{Evaluation of Inputs}

\begin{figure}[ht!]
\centering
\includegraphics[width=1\linewidth]{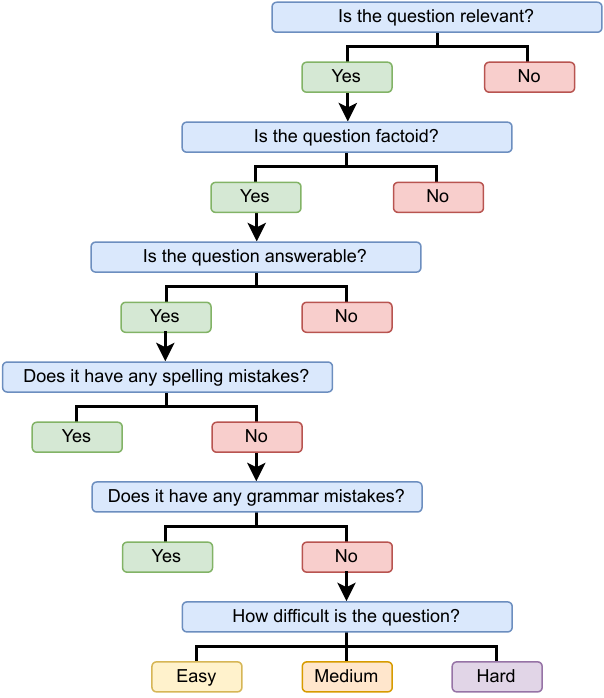}
\caption{Hierarchical evaluation of the questions.}
\label{fig:question}
\end{figure}

Figure~\ref{fig:question} shows a set of evaluation criteria for evaluating the MRC questions. The question is \textit{relevant} if it is on the topic covered in the corresponding text corpus. \textit{Factoid} questions are questions that start with one of the following words: ``who'', ``what'', ``where'', ``when'', ``why'' or ``how''. They ask about facts that can be expressed as short texts \cite{parsing2009speech}. The question is \textit{answerable} if there exists an answer to it. The evaluators are asked if the posed question contains any \textit{spelling} or \textit{grammar} errors. The \textit{difficulty} of the posed question can be chosen from three levels – \textit{easy}, \textit{medium}, or \textit{hard} (please refer to Table~\ref{tab:difficulty}).

\begin{table}[!ht]
\centering
\caption{Three different levels of difficulty of the posed questions.} 
\begin{tabular}{|c|l|}
\hline
Easy & \makecell{The correct answer is \\ obvious after reading the passage \\ only one time.} \\ 
\hline
Medium & \makecell{To find the correct answer, one \\ needs to carefully read and \\ understand both the question and \\ the paragraph.}  \\ 
\hline
Hard & \makecell{To find the correct answer, \\ one needs to read the paragraph \\ many times, sometimes even use \\ logical reasoning to find the \\ correct answer.} \\ 
\hline
\end{tabular}
\label{tab:difficulty}
\end{table}

\subsubsection{Evaluation of Outputs}

The evaluators were asked to evaluate the retrieved \textit{short answer} and if necessary its \textit{explanation}. Namely, the output of the whole MRC system is a text span (i.e., short answer). However, an MRC system can be seen as a pipeline of two NLP models - \textit{document retrieval} and \textit{document reader}, where the output of the former model is the \textit{relevant passage(s)} and the output of the latter model (i.e., the whole system) is a \textit{text span}. Our metric first evaluates the characteristics of the output of the whole system (i.e., text span). If the output of the whole system was not satisfying, then we evaluate its explanation (i.e., relevant passage) that was provided by the document retrieval component. 

\begin{figure*}[t!]
\centering
\includegraphics[width=0.96\linewidth]{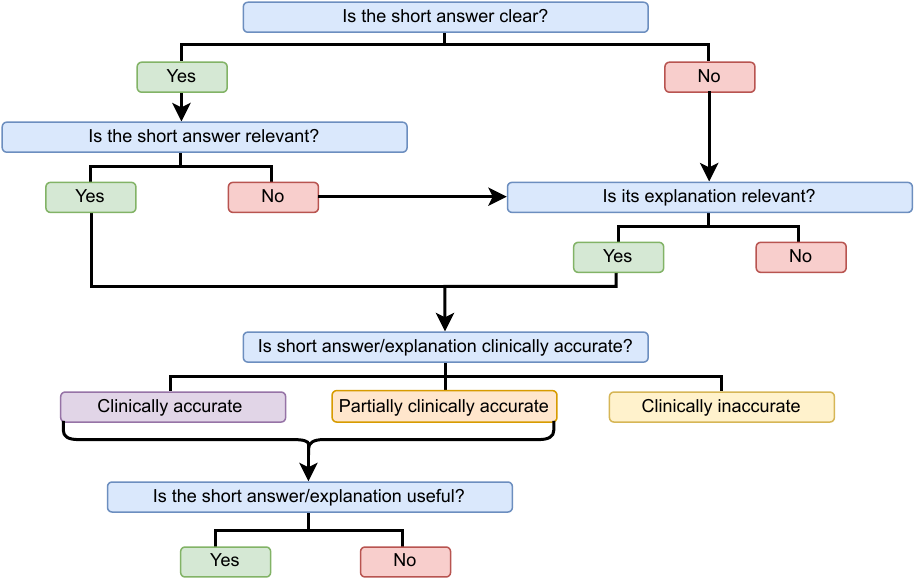}
\caption{Hierarchical evaluation of the answers.}
\label{fig:answer}
\end{figure*}

The retrieved short answer is \textit{clear} if its meaning is easy to understand. The retrieved short answer/explanation is \textit{relevant} if it answers the posed question. \textit{Clinical accuracy} of the retrieved short answer/explanation denotes the degree to which it is clinically accurate – (i) clinically accurate, (ii) partially clinically accurate, and (iii) clinically inaccurate (see Table \ref{tab:accuracy}). Finally, the health coaches judged the usefulness of the retrieved short answer/explanation (see Figure \ref{fig:answer}).

\begin{table}[!ht]
\centering
\caption{Three different levels of clinical accuracy.} 
\begin{tabular}{|c|l|}
\hline
\makecell{Clinically \\ accurate} & \makecell{The retrieved short answer/ \\ explanation is clinically accurate \\ and is based on evidence-based \\ information.} \\ 
\hline
\makecell{Partially \\ clinically \\ accurate} & \makecell{The retrieved short answer/ \\ explanation is partially clinically \\ accurate and somewhat lacks \\ evidence-based information.}  \\ 
\hline
\makecell{Clinically \\ inaccurate} & \makecell{The retrieved short answer/ \\ explanation is not clinically \\ accurate and is not based \\ on evidence-based information.} \\ 
\hline
\end{tabular}
\label{tab:accuracy}
\end{table}

\subsection{Testing of the MRC System}

Testing of the developed MRC system was conducted during a pilot Randomized Controlled Trial (RCT). In this RCT, 30 participants in the intervention group (i.e., clients) interacted with 10 health coaches who utilized the MRC system to answer factoid questions. Clients were recruited from a general student population if they (1) were older than 21 years, (2) were available for weekly interaction with a health coach for four weeks, (3) were not currently undergoing any treatment for a sleep disorder or mental disorder and were not under the care of a psychologist or psychiatrist, and (iv) had PHQ-9 score less than 10. 

Health coaches were recruited from the cohorts of graduated students from the health coaching course if they (1) were older than 21 years, (2) were available for weekly interaction with three clients for four weeks, and (iii) successfully completed and passed the health coaching course. During the study period of four weeks, clients had weekly 30-minute sessions with their respective health coaches. All questions asked by health coaches and their corresponding answers were saved during the testing phase and were subsequently used in the evaluation phase. By dividing human evaluation into two parts, we were able also to judge whether questions were posed in the way we asked our health coaches to ask them, i.e., if they can be answered by the developed MRC system.

\subsection{Evaluation of the MRC System}

\begin{figure}[ht!]
\centering
\includegraphics[width=1\linewidth]{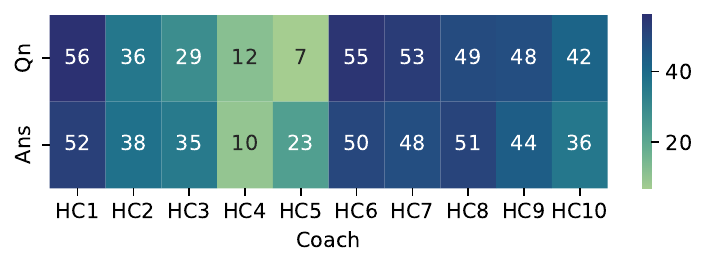}
\caption{The total number of questions and answers evaluated by each health coach.}
\label{fig:count}
\end{figure}

\begin{figure}[ht!]
\centering
\includegraphics[width=1\linewidth]{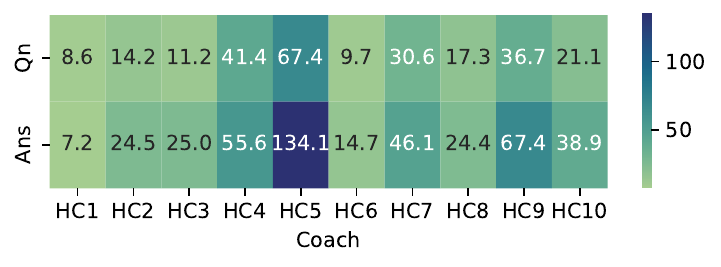}
\caption{Average time in seconds per health coach needed to evaluate questions and answers.}
\label{fig:time}
\end{figure}

Following a 4-week pilot RCT, the developed MRC system underwent evaluation by 10 health coaches. A total of 387 unique question-answer pairs were evaluated by the health coaches during this period. The heat map depicted in Figure \ref{fig:count} illustrates the number of inputs and outputs evaluated by each health coach, while Figure \ref{fig:time} showcases the average evaluation time required for each input/output assessed by the health coaches.

\begin{figure}[ht!]
\centering
\includegraphics[width=1\linewidth]{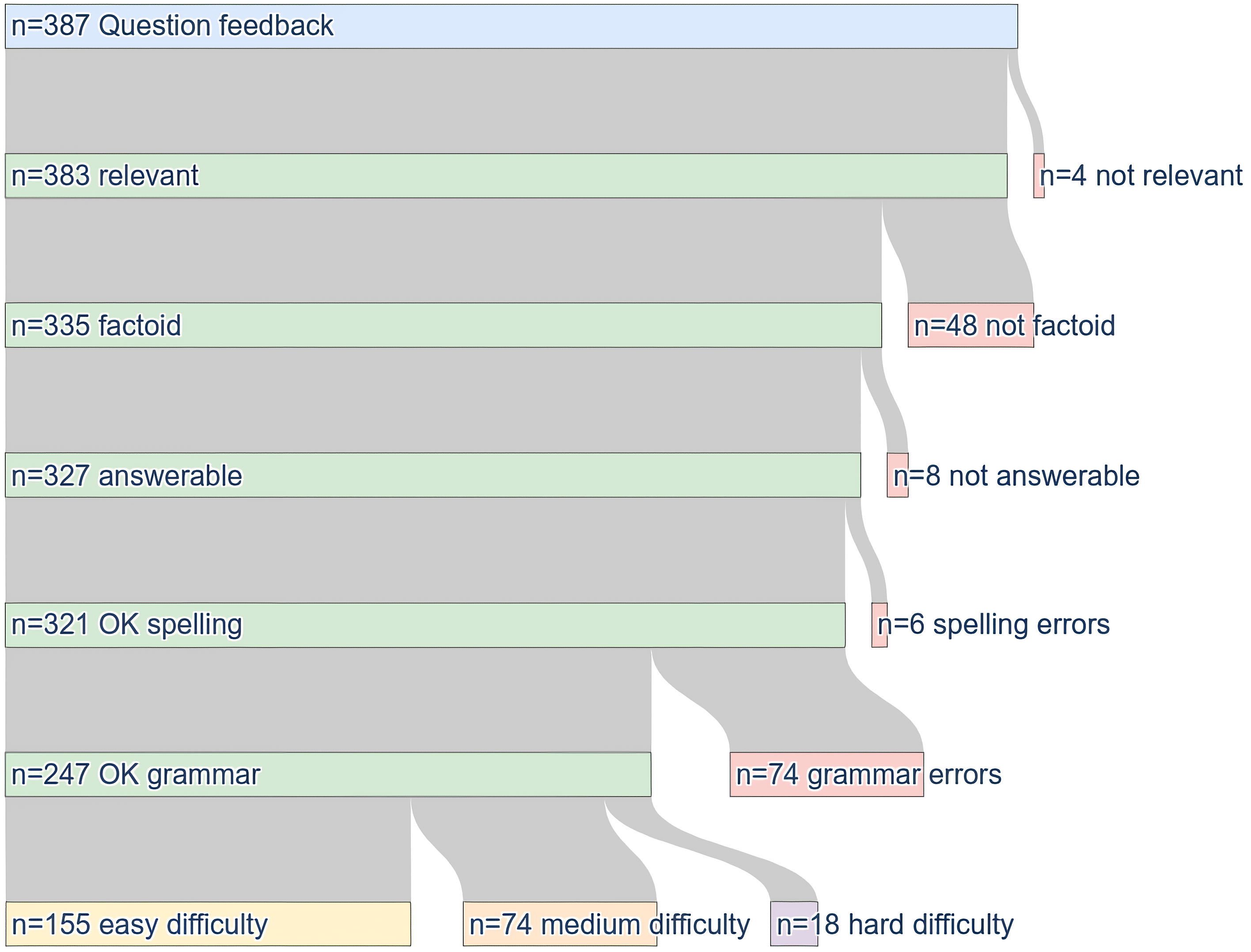}
\caption{Extrinsic evaluation of questions.}
\label{fig:ee_q}
\end{figure}

Almost all questions (99\%, 383/387) were evaluated as \textit{relevant}. One example of a question that was marked as not relevant was: "\textit{Food nutrition tips}". The next 87\% of questions (335/383) were judged as factoid. Some examples of not factoid questions are as follows: "\textit{About REM sleep, is it the phase that I'm dreaming?}", "\textit{Can you exercise before sleeping?}", "\textit{I often run around campus for 3-5km at night 1-2h before sleeping. Is it good or bad for sleep?}". 2\% of the remaining questions (8/335) were marked as not answerable: \textit{"How long should I be awake during sleep?"}, \textit{"How bad would you say is my sleep health like compared to the average?"}, while additional 2\% (6/327) had spelling errors (e.g., \textit{"How long before bedtime shld i stop screentime?"}). Finally, the last 23\% (74/321) had grammar errors: \textit{"How do ensure naps have good quality?"}, \textit{"Why wake up during night?"}. The results of the complete external human evaluation for questions are shown in Figure \ref{fig:ee_q}.

\begin{figure}[ht!]
\centering
\includegraphics[width=1\linewidth]{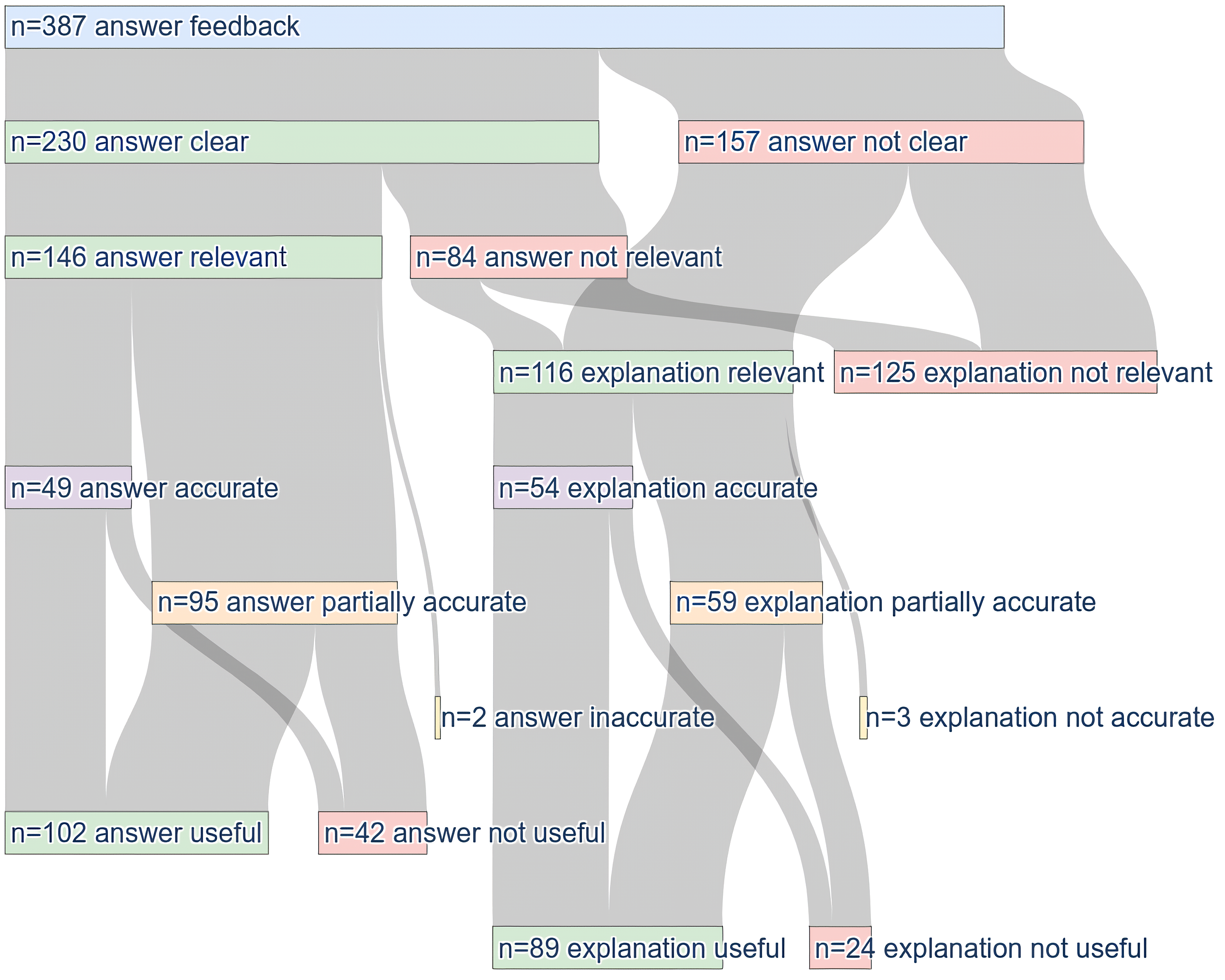}
\caption{Extrinsic evaluation of answers.}
\label{fig:ee_a}
\end{figure}

More than 40\% (157/387) of short answers were evaluated as not \textit{clear}, out of which in 57\% of cases (89/157), their explanations were marked as relevant. For example, "\textit{\textbf{Question}: When does melatonin peak? \textbf{Answer}: release of melatonin, the hormone that induces feelings of tiredness and relaxation. \textbf{Explanation}: When the sun goes down, your eyes will perceive darkness and signal the scn accordingly. This triggers the release of melatonin, the hormone that induces feelings of tiredness and relaxation. This also causes your core temperature to dip.}". 63\% of clear answers (146/230) were also evaluated as relevant of which 99\% (144/146) was indicated as being (partly) clinically accurate. Furthermore, 97\% (113/116) of the short answers that were not clear, but their explanations were relevant, were (partly) clinically accurate. The results of the complete external human evaluation for answers are shown in Figure \ref{fig:ee_a}.

\subsection{Composite scores of the MRC System}

The results of our evaluation showed that 63.8\% (247/387) of unique questions were evaluated as relevant, factoid, answerable, spelling and grammar mistakes-free (i.e., \textit{good} questions). Out of those, 63\% (155/247) were judged as easy, 30\% (74/247) as medium and 7\% (18/247) as hard questions. Furthermore, 49.4\% (191/387) of unique answers were evaluated as clear, relevant, clinically accurate and useful (i.e., \textit{good} answers). In order to check if there are any associations between the quality of outputs and inputs, we performed a $\chi^2$ test. The result showed significant associations between the two ($\chi^2$ = 4.56, p=0.03). The distribution of the performance matrix is shown in Table~\ref{tab:confusion}.

\begin{table}[ht!]
\centering
\caption{2x2 matrix for the performed $\chi^2$ test.} 
\begin{tabular}{|cc|cc|l}
\cline{1-4}
\multicolumn{2}{|c|}{\multirow{2}{*}{}}               & \multicolumn{2}{c|}{Questions}  &  \\ \cline{3-4}
\multicolumn{2}{|c|}{}                                & \multicolumn{1}{c|}{good} & bad &  \\ \cline{1-4}
\multicolumn{1}{|c|}{\multirow{2}{*}{Answers}} & good & \multicolumn{1}{c|}{132}  & 59  &  \\ \cline{2-4}
\multicolumn{1}{|c|}{}                         & bad  & \multicolumn{1}{c|}{115}  & 81  &  \\ \cline{1-4}
\end{tabular}
\label{tab:confusion}
\end{table}

%% file: Sections/5_dis.tex
In this study, we conducted a scoping review to identify gaps in the literature regarding human evaluation in NLP. 
%Our review focused on various aspects of conducting human evaluation in NLP tasks, including evaluator characteristics, evaluation samples, scoring systems, evaluation designs, and statistical analysis. 
The findings revealed three significant gaps that need to be addressed: the lack of evaluation metrics for NLP system inputs, limited consideration for interdependencies among different characteristics of NLP systems, and a scarcity of metrics for extrinsic evaluation.

To bridge these gaps and enhance human evaluation in NLP, we proposed a hierarchical evaluation framework. Our framework offers a standardized approach that considers both the inputs and outputs of NLP systems, allowing for a more comprehensive assessment. 
%By incorporating the evaluation of inputs, we aim to evaluate the quality and relevance of the data provided to NLP systems, which is often overlooked in current evaluation practices. 
Moreover, our hierarchical approach considers the interdependencies among different characteristics of NLP systems. Rather than evaluating characteristics independently, our framework emphasizes their interconnectedness and the impact they may have on each other. This approach enables a more holistic evaluation that captures the overall performance of NLP systems.

To validate the effectiveness of our proposed framework, we conducted a pilot RCT evaluating an MRC system. The evaluation phase of our study involved 10 health coaches who evaluated a total of 387 question-answer pairs generated during the RCT. The evaluation metrics developed for inputs focused on aspects such as relevance, factoid nature, answerability, spelling, grammar errors, and difficulty levels of the questions. For outputs, the evaluation criteria included clarity, relevance, clinical accuracy, and usefulness of the retrieved short answers and explanations.

The results of the evaluation provided valuable insights into the strengths and weaknesses of the MRC system and demonstrated the practical application of our hierarchical evaluation framework. The findings supported the notion that evaluating both inputs and outputs is crucial for obtaining a comprehensive understanding of the performance and effectiveness of NLP systems. Future research should focus on validating the scalability and time-saving benefits of our proposed framework.

%% file: paper.bbl
\begin{thebibliography}{26}
\expandafter\ifx\csname natexlab\endcsname\relax\def\natexlab#1{#1}\fi

\bibitem[{Amidei et~al.(2018{\natexlab{a}})Amidei, Piwek, and
  Willis}]{amidei2018evaluation}
Jacopo Amidei, Paul Piwek, and Alistair Willis. 2018{\natexlab{a}}.
\newblock Evaluation methodologies in automatic question generation 2013-2018.
\newblock In \emph{Proceedings of the 11th International Conference on Natural
  Language Generation}.

\bibitem[{Amidei et~al.(2018{\natexlab{b}})Amidei, Piwek, and
  Willis}]{amidei2018rethinking}
Jacopo Amidei, Paul Piwek, and Alistair Willis. 2018{\natexlab{b}}.
\newblock Rethinking the agreement in human evaluation tasks.
\newblock In \emph{Proceedings of the 27th International Conference on
  Computational Linguistics}, pages 3318--3329.

\bibitem[{Belz and Gatt(2008)}]{belz2008intrinsic}
Anja Belz and Albert Gatt. 2008.
\newblock Intrinsic vs. extrinsic evaluation measures for referring expression
  generation.
\newblock In \emph{Proceedings of ACL-08: HLT, Short Papers}, pages 197--200.

\bibitem[{Bojic et~al.(2023{\natexlab{a}})Bojic, Halim, Suharman, Tar, Ong,
  Phung, Ravaut, Joty, and Car}]{bojic-etal-2023-data}
Iva Bojic, Josef Halim, Verena Suharman, Sreeja Tar, Qi~Chwen Ong, Duy Phung,
  Mathieu Ravaut, Shafiq Joty, and Josip Car. 2023{\natexlab{a}}.
\newblock \href {https://aclanthology.org/2023.insights-1.3} {A data-centric
  framework for improving domain-specific machine reading comprehension
  datasets}.
\newblock In \emph{The Fourth Workshop on Insights from Negative Results in
  NLP}, pages 19--32, Dubrovnik, Croatia. Association for Computational
  Linguistics.

\bibitem[{Bojic et~al.(2023{\natexlab{b}})Bojic, Ong, Joty, and
  Car}]{bojic2023building}
Iva Bojic, Qi~Chwen Ong, Shafiq Joty, and Josip Car. 2023{\natexlab{b}}.
\newblock Building extractive question answering system to support human-ai
  health coaching model for sleep domain.
\newblock \emph{arXiv preprint arXiv:2305.19707}.

\bibitem[{Bojic et~al.(2022)Bojic, Ong, Thakkar, Kamran, Le~Shua, Pang, Chen,
  Nayak, Joty, and Car}]{bojic2022sleepqa}
Iva Bojic, Qi~Chwen Ong, Megh Thakkar, Esha Kamran, Irving~Yu Le~Shua, Jaime
  Rei~Ern Pang, Jessica Chen, Vaaruni Nayak, Shafiq Joty, and Josip Car. 2022.
\newblock Sleepqa: A health coaching dataset on sleep for extractive question
  answering.
\newblock In \emph{Machine Learning for Health}, pages 199--217. PMLR.

\bibitem[{Castilho(2021)}]{castilho2021towards}
Sheila Castilho. 2021.
\newblock Towards document-level human mt evaluation: On the issues of
  annotator agreement, effort and misevaluation.
\newblock In \emph{Proceedings of the Workshop on Human Evaluation of NLP
  Systems (HumEval)}. Association for Computational Linguistics (ACL).

\bibitem[{Gangal et~al.(2022)Gangal, Feng, Alikhani, Mitamura, and
  Hovy}]{gangal2022nareor}
Varun Gangal, Steven~Y Feng, Malihe Alikhani, Teruko Mitamura, and Eduard Hovy.
  2022.
\newblock Nareor: The narrative reordering problem.
\newblock In \emph{Proceedings of the AAAI Conference on Artificial
  Intelligence}, pages 10645--10653.

\bibitem[{Gehrmann et~al.(2023)Gehrmann, Clark, and
  Sellam}]{gehrmann2023repairing}
Sebastian Gehrmann, Elizabeth Clark, and Thibault Sellam. 2023.
\newblock Repairing the cracked foundation: A survey of obstacles in evaluation
  practices for generated text.
\newblock \emph{Journal of Artificial Intelligence Research}, 77:103--166.

\bibitem[{Howcroft et~al.(2020)Howcroft, Belz, Clinciu, Gkatzia, Hasan,
  Mahamood, Mille, Van~Miltenburg, Santhanam, and Rieser}]{howcroft2020twenty}
David Howcroft, Anya Belz, Miruna Clinciu, Dimitra Gkatzia, Sadid~A Hasan, Saad
  Mahamood, Simon Mille, Emiel Van~Miltenburg, Sashank Santhanam, and Verena
  Rieser. 2020.
\newblock Twenty years of confusion in human evaluation: Nlg needs evaluation
  sheets and standardised definition.
\newblock In \emph{Proceedings of the 13th International Conference on Natural
  Language Generation}. Association for Computational Linguistics (ACL).

\bibitem[{Iskender et~al.(2021)Iskender, Polzehl, and
  M{\"o}ller}]{iskender2021reliability}
Neslihan Iskender, Tim Polzehl, and Sebastian M{\"o}ller. 2021.
\newblock Reliability of human evaluation for text summarization: Lessons
  learned and challenges ahead.
\newblock In \emph{Proceedings of the Workshop on Human Evaluation of NLP
  Systems (HumEval)}, pages 86--96.

\bibitem[{Kurisinkel and Chen(2019)}]{kurisinkel2019set}
Litton~J Kurisinkel and Nancy Chen. 2019.
\newblock Set to ordered text: Generating discharge instructions from medical
  billing codes.
\newblock In \emph{Proceedings of the 2019 Conference on Empirical Methods in
  Natural Language Processing and the 9th International Joint Conference on
  Natural Language Processing (EMNLP-IJCNLP)}, pages 6165--6175.

\bibitem[{Liang and Li(2021)}]{liang2021towards}
Hongru Liang and Huaqing Li. 2021.
\newblock Towards standard criteria for human evaluation of chatbots: A survey.
\newblock \emph{arXiv preprint arXiv:2105.11197}.

\bibitem[{Liu et~al.(2022)Liu, Fabbri, Liu, Zhao, Nan, Han, Han, Joty, Wu,
  Xiong et~al.}]{liu2022revisiting}
Yixin Liu, Alexander~R Fabbri, Pengfei Liu, Yilun Zhao, Linyong Nan, Ruilin
  Han, Simeng Han, Shafiq Joty, Chien-Sheng Wu, Caiming Xiong, et~al. 2022.
\newblock Revisiting the gold standard: Grounding summarization evaluation with
  robust human evaluation.
\newblock \emph{arXiv preprint arXiv:2212.07981}.

\bibitem[{Otani et~al.(2023)Otani, Togashi, Sawai, Ishigami, Nakashima, Rahtu,
  Heikkil{\"a}, and Satoh}]{otani2023toward}
Mayu Otani, Riku Togashi, Yu~Sawai, Ryosuke Ishigami, Yuta Nakashima, Esa
  Rahtu, Janne Heikkil{\"a}, and Shin’ichi Satoh. 2023.
\newblock Toward verifiable and reproducible human evaluation for text-to-image
  generation.
\newblock In \emph{Proceedings of the IEEE/CVF Conference on Computer Vision
  and Pattern Recognition}, pages 14277--14286.

\bibitem[{Paroubek et~al.(2007)Paroubek, Chaudiron, and
  Hirschman}]{paroubek-etal-2007-principles}
Patrick Paroubek, St{\'e}phane Chaudiron, and Lynette Hirschman. 2007.
\newblock \href {https://aclanthology.org/2007.tal-1.1} {Principles of
  evaluation in natural language processing}.
\newblock In \emph{Traitement Automatique des Langues, Volume 48, Num{\'e}ro 1
  : Principes de l'{\'e}valuation en Traitement Automatique des Langues
  [Principles of Evaluation in Natural Language Processing]}, pages 7--31,
  France. ATALA (Association pour le Traitement Automatique des Langues).

\bibitem[{Parsing(2009)}]{parsing2009speech}
Constituency Parsing. 2009.
\newblock Speech and language processing.
\newblock \emph{Power Point Slides}.

\bibitem[{Peters et~al.(2015)Peters, Godfrey, McInerney, Soares, Khalil, and
  Parker}]{peters2015joanna}
Micah~DJ Peters, Christina~M Godfrey, Patricia McInerney, Cassia~Baldini
  Soares, Hanan Khalil, and Deborah Parker. 2015.
\newblock \emph{The Joanna Briggs Institute reviewers' manual 2015: methodology
  for JBI scoping reviews}, chapter~11. The Joanna Briggs Institute.

\bibitem[{Qian and Levy(2022)}]{qian2022flexible}
Peng Qian and Roger Levy. 2022.
\newblock Flexible generation from fragmentary linguistic input.
\newblock In \emph{Proceedings of the 60th Annual Meeting of the Association
  for Computational Linguistics (Volume 1: Long Papers)}, pages 8176--8196.

\bibitem[{Son et~al.(2022)Son, Jin, Yoo, Bak, Cho, and Oh}]{son2022translating}
Juhee Son, Jiho Jin, Haneul Yoo, JinYeong Bak, Kyunghyun Cho, and Alice Oh.
  2022.
\newblock Translating hanja historical documents to contemporary korean and
  english.
\newblock In \emph{Findings of the Association for Computational Linguistics:
  EMNLP 2022}, pages 1260--1272.

\bibitem[{Sudoh et~al.(2021)Sudoh, Takahashi, and
  Nakamura}]{sudoh2021translation}
Katsuhito Sudoh, Kosuke Takahashi, and Satoshi Nakamura. 2021.
\newblock Is this translation error critical?: Classification-based human and
  automatic machine translation evaluation focusing on critical errors.
\newblock In \emph{Proceedings of the Workshop on Human Evaluation of NLP
  Systems (HumEval)}, pages 46--55.

\bibitem[{van~der Lee et~al.(2021)van~der Lee, Gatt, van Miltenburg, and
  Krahmer}]{van2021human}
Chris van~der Lee, Albert Gatt, Emiel van Miltenburg, and Emiel Krahmer. 2021.
\newblock Human evaluation of automatically generated text: Current trends and
  best practice guidelines.
\newblock \emph{Computer Speech \& Language}, 67:101151.

\bibitem[{Van Der~Lee et~al.(2019)Van Der~Lee, Gatt, Van~Miltenburg, Wubben,
  and Krahmer}]{van2019best}
Chris Van Der~Lee, Albert Gatt, Emiel Van~Miltenburg, Sander Wubben, and Emiel
  Krahmer. 2019.
\newblock Best practices for the human evaluation of automatically generated
  text.
\newblock In \emph{Proceedings of the 12th International Conference on Natural
  Language Generation}, pages 355--368.

\bibitem[{Zeng and Nie(2021)}]{zeng2021investigation}
Yan Zeng and Jian-Yun Nie. 2021.
\newblock An investigation of suitability of pre-trained language models for
  dialogue generation--avoiding discrepancies.
\newblock In \emph{Findings of the Association for Computational Linguistics:
  ACL-IJCNLP 2021}, pages 4481--4494.

\bibitem[{Zhang et~al.(2020)Zhang, Duckworth, Ippolito, and
  Neelakantan}]{zhang2020trading}
Hugh Zhang, Daniel Duckworth, Daphne Ippolito, and Arvind Neelakantan. 2020.
\newblock Trading off diversity and quality in natural language generation.
\newblock \emph{arXiv preprint arXiv:2004.10450}.

\bibitem[{Zhu et~al.(2020)Zhu, Zhang, Liu, and Wang}]{zhu2020counterfactual}
Qingfu Zhu, Weinan Zhang, Ting Liu, and William~Yang Wang. 2020.
\newblock Counterfactual off-policy training for neural dialogue generation.
\newblock In \emph{Proceedings of the 2020 Conference on Empirical Methods in
  Natural Language Processing (EMNLP)}, pages 3438--3448.

\end{thebibliography}
